\documentclass[sort&compress, numafflabel]{elsarticle}

\usepackage[]{natbib}
\usepackage[breaklinks,hidelinks]{hyperref}
\usepackage{times}
\usepackage{latexsym}

\usepackage{geometry}
\usepackage{subfiles}
\usepackage{multirow}
\usepackage{array}
\usepackage{hyphenat}
\usepackage{subcaption}
\usepackage{longtable}
\usepackage{graphicx}
\usepackage{adjustbox}
\usepackage{enumitem}
\usepackage{url}

\usepackage{bm}
\usepackage{booktabs}
\usepackage{float}
\usepackage[english]{babel}
\usepackage{blindtext}
\usepackage{textcomp}
\usepackage{soul}
\usepackage{pifont}

\makeatletter
\def\ps@pprintTitle{%
 \let\@oddhead\@empty
 \let\@evenhead\@empty
 \def\@oddfoot{}%
 \let\@evenfoot\@oddfoot}
\makeatother
\usepackage{epstopdf}
\AppendGraphicsExtensions{.tif}

\usepackage{titlesec}
\titleformat{\section}
      {\normalfont\bfseries}
      {\thesection}
      {0ex}
      {\MakeUppercase}

\titleformat{\subsection}
      {\normalfont\bfseries}
      {\thesection}
      {0ex}
      {}

\titleformat{\subsubsection}
      {\normalfont}
      {\thesection}
      {0ex}
      {}

\newcommand\blfootnote[1]{%
  \begingroup
  \renewcommand\thefootnote{}\footnote{#1}%
  \addtocounter{footnote}{-1}%
  \endgroup
}

\usepackage{setspace}
\newif\ifsubfile
\subfiletrue
\newif\iftif
\tiffalse

\usepackage[utf8]{inputenc}
\usepackage[T1]{fontenc}
\usepackage{lineno}
\usepackage[numbers]{natbib}
\usepackage{subfiles}
\usepackage{graphicx}
\usepackage{float}
\usepackage{hyperref}
\usepackage{mathtools}
\usepackage{changepage}
\usepackage{caption}
\usepackage{multirow}
\usepackage{titlesec}
\usepackage{amsmath}

\usepackage{makecell}

\usepackage[math]{cellspace}
\usepackage{tikz}

\usepackage{url}

\usepackage{colortbl}
\linenumbers
\nolinenumbers

\usepackage{listings}

\title{Estimating Outpatient Visit Prices from Hospital Price Transparency Data using Large Language Models}

\author[inst1]{Kurmanbek Kaiyrbekov, PhD \textsuperscript{* ,} \blfootnote{\textsuperscript{*} Corresponding author: Kurmanbek Kaiyrbekov, PhD, Center for Genomics and Data Science Research, National Human Genome Research Institute, National Institutes of Health, Bethesda, Maryland, USA; kurmanbek.kaiyrbekov@nih.gov}}

\ead{email@example.edu}
\author[inst2]{Nicholas J Dobbins, PhD, MLIS}\ead{}
\author[inst3]{Sean D Mooney, PhD}\ead{}

\address[inst1]{Center for Genomics and Data Science Research, National Human Genome Research Institute, National Institutes of Health, Bethesda, Maryland, USA}
\address[inst2]{Office of Intramural Research, National Institutes of Health, Bethesda, Maryland, USA
\blfootnote{ \\ Word count: 3999} }

\begin{document}
\subfilefalse

\newpageafter{author}

\begin{abstract}

\noindent\textbf{Objective:}  

Hospital care prices in the United States have increased substantially, yet patients lack effective tools to access and compare costs. We aimed to develop and evaluate an LLM-based approach for improving access to and use of hospital price transparency (PT) data.  
  
\noindent\textbf{Materials and Methods:}    

To estimate visit prices, we used a retrieval-augmented generation (RAG) framework to analyze PT files from three major hospitals in Maryland and the surrounding region. We compared BM25, OpenAI embedding models, and a GPT--BM25 hybrid against a no-RAG baseline for cost estimation of individual procedures and procedure groups, then evaluated the best approach for symptom-based visit cost estimation using synthetic simulations.
  
\noindent\textbf{Results:}  

Across hospitals, the GPT--BM25 retriever achieved the best performance, with average price MAPE of 19.44\% for individual procedures and 4.90\% for groups, compared with 147.98\% and 82.82\%, respectively, for the no-RAG baseline. In simulations, our framework correctly identified approximately 50\% of reference procedures and reduced MAPE relative to the LLM-only baseline at two of the three hospitals, while achieving comparable performance at the third.
  
\noindent\textbf{Discussion:}  

Our framework uses RAG to improve care price estimation in real world settings. Compared with the no-RAG baseline, it substantially improved cost estimates for both individual procedures and procedure groups. Improvements were smaller in synthetic simulations, likely because inferring procedures from clinical information is more complex. Practical challenges remain, including heterogeneous PT file formats, unexplained duplicate entries, and limited availability of insurer-negotiated rates.

\noindent\textbf{Conclusion:}  

LLMs that effectively integrate PT data show strong potential to provide patients with more accessible and accurate healthcare cost estimates.

\end{abstract}

\maketitle

\pagebreak

\section*{Introduction}
\label{sec:background}

Large language models (LLMs) have rapidly expanded the scope of artificial intelligence in healthcare, with emerging applications in diagnosis, clinical documentation and summarization, clinical decision support, and medical coding \cite{singhal2023large, mcduff2025towards, ferber2026towards, abacha2023empirical, hou2025enhancing, kim2024mdagents}. These models have the potential to improve the quality of care by reducing the administrative and cognitive burden on physicians and other healthcare professionals. However, in addition to quality of care, affordability remains a central concern for patients in the United States, particularly as out-of-pocket healthcare spending increases rapidly \cite{amin2025healthcare, gaffney2026risk}. Although increasing hospital prices are driven by multiple factors, a major contributor is the persistent lack of transparency in prices, which limits patients’ ability to compare options and seek lower-cost care \cite{chang2026high}. To address this problem, in November 2019, the Centers for Medicare and Medicaid Services (CMS) issued a rule requiring hospitals to publish price transparency files containing prices for “shoppable” services (services that can be scheduled by a healthcare consumer in advance) beginning January 1, 2021 \cite{cms_hpt}. The rule was intended to make it easier for consumers to compare prices across hospitals and estimate the cost of care before receiving services. However, while these data are technically machine-readable, they are often released without a standardized structure or format, making them difficult to analyze and interpret and thereby limiting their usefulness for meaningful price comparison \cite{maleki2024usability}.  In this work, we aim to improve healthcare price transparency by developing an LLM-based approach that facilitates access to, interpretation of, and comparison of hospital price-transparency data.

\section*{Background and Significance}
\label{sec:background}

Hospital prices in the United States vary widely across regions, health systems, and payers.  Prior studies have documented large differences in commercial negotiated prices for common inpatient and outpatient services across states, metropolitan areas, and within states \cite{bai2015extreme, cooper2019price, linde2022hospital} due to differences in provider market concentration, insurer competition, local practice patterns, and regulatory environments. Larger health systems and hospitals with greater market power often command higher prices than smaller or less consolidated institutions \cite{white2014understanding}, even when differences in service quality are limited \cite{roberts2017high, hussey2013association}. In addition, services delivered in physician offices are frequently cheaper than services provided in hospital outpatient departments \cite{sen2022site}. Negotiated rates also vary across payers, with larger insurers often obtaining lower service prices than smaller insurers at the same hospital, likely because of greater bargaining leverage \cite{wang2024within, wang2024insurers}. So, hospital pricing is highly heterogeneous and shaped by a complex interplay of provider characteristics, negotiating power, and regional market conditions. 

Wide variation in hospital prices could allow patients to lower spending by choosing lower-cost providers, but only if price information is accessible and interpretable when care decisions are made. Although CMS requires hospitals to disclose prices publicly, these files are often fragmented, inconsistently structured, and difficult for consumers to use. Hence, most previous studies using hospital price transparency data have focused on hospital compliance, market effects, policy evaluation, and recommendations to strengthen regulation and reporting standards \cite{sen2022site, wang2024within}. Although this work has generated important insights, a central practical question remains unresolved: can currently available hospital price transparency files be translated into tools that support patient price comparison at the time care decisions are made? To explore this question, we prototyped an LLM-based framework using price transparency files from three hospitals. We assess whether our framework can interpret heterogeneous data formats, identify relevant services and associated prices, generate cross-hospital price comparisons, and approximate the cost of an outpatient visit in advance of care. We focus on outpatient visits because most services reported in hospital price transparency files are delivered in outpatient settings.

\section*{Materials and Methods}
\label{sec:methods}

\subsection*{Data}  
  
Because patients typically receive care within their local geographic region, we evaluated our approach using hospital price transparency files from three large hospitals in Maryland and the surrounding area: Johns Hopkins Hospital (Baltimore, MD) \cite{hopkinsmedicine_charges_fees}, the University of Maryland Medical Center \cite{umms_price_transparency}, Downtown Campus (Baltimore, MD), and MedStar Washington Hospital Center (Washington, DC) \cite{medstar_price_transparency}. For brevity, we refer to these institutions as Hopkins, UMaryland, and MedStar, respectively.  
  
The PT files were provided in comma-separated value format and contained abbreviated service descriptions, associated billing codes, and multiple pricing fields. These included gross charges (standard list prices from the chargemaster), discounted cash prices (reduced rates offered to self-paying patients), and payer-specific negotiated charges (amounts negotiated between the hospital and individual insurers).

\subsection*{Framework architecture}  
Hospital PT files typically report prices at the level of billing records, such as CPT-coded procedures, rather than patient-friendly descriptions of care. For example, a patient presenting with wrist pain after a fall may ask about the cost of a "wrist x-ray", but the relevant pricing records in a hospital PT file may correspond to one or more specific CPT codes for radiologic examination of the wrist, depending on the number of views obtained and how the service is coded. The task therefore involves identifying which CPT-coded records most likely correspond to the patient's requested service before retrieving the associated prices.  
  
\begin{figure}
    \centering
    \includegraphics[width=0.95\linewidth]{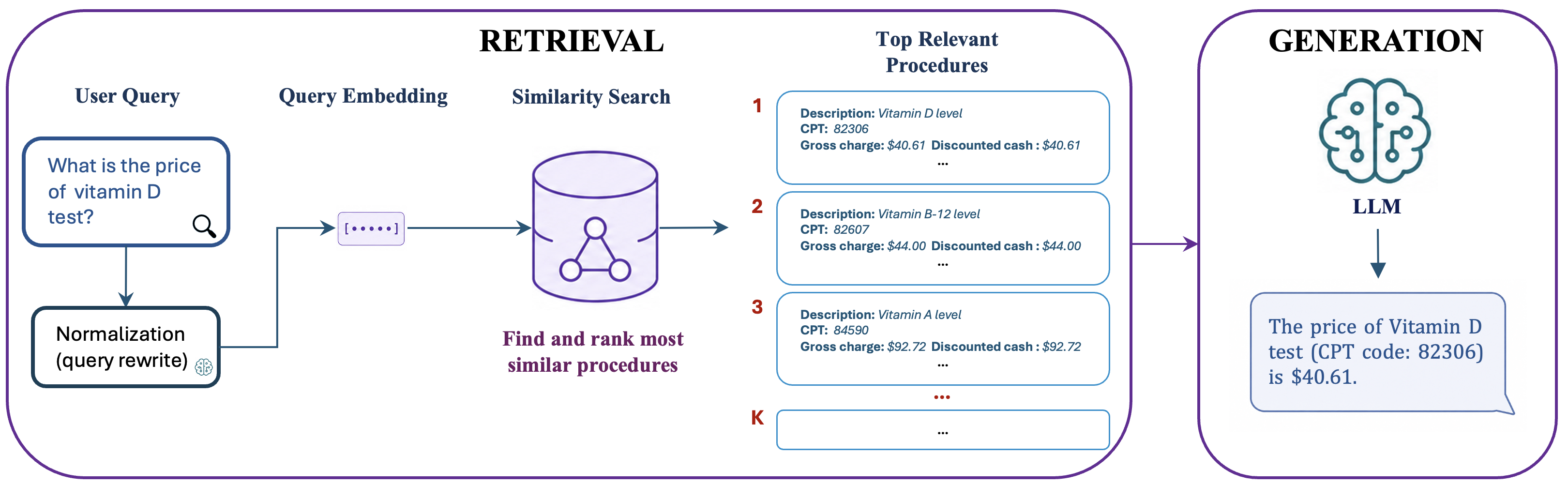}
    \caption{Overview of the RAG framework for grounding pricing responses in hospital transparency files. The retrieval stage normalizes and embeds the user query, then performs a similarity search against a vector database to identify the top-$K$ candidate records. In the generation stage, an LLM selects the procedure and price that best match the original query from among the retrieved records.}
    \label{fig:rag_framework}
\end{figure}

Accurate price identification from PT files requires both reliable retrieval of relevant records and correct interpretation of those records. To address these challenges, we used a RAG architecture \cite{lewis2020retrieval} with hospital PT files as the primary data source. As shown in Figure~\ref{fig:rag_framework}, a user query is first rewritten by the LLM to better align with official procedure descriptions and account for variation in phrasing. The rewritten query is then embedded and used to retrieve candidate procedures from a vector database via approximate nearest neighbor search \cite{douze2025faiss} with cosine similarity. The LLM then selects the best-matching procedure and generates a response grounded in hospital-specific pricing data rather than unsupported estimates. We used OpenAI's gpt-5.4 (version: 2026-03-05) in all evaluations unless otherwise noted.

We evaluated four retrieval strategies representing sparse, dense, and hybrid approaches:  
  
\begin{itemize}  
    \item \textbf{Okapi BM25.} BM25 is a probabilistic sparse-retrieval algorithm that ranks records according to lexical term matching. It incorporates term frequency, inverse document frequency, document-length normalization, and diminishing returns for repeated terms \cite{whissell2011improving}.  
  
    \item \textbf{OpenAI text-embedding-3-large.} This model maps text to dense vectors in a 3,072-dimensional embedding space \cite{openai_embeddings}. These embeddings capture semantic similarity and can retrieve records that describe similar clinical procedures using different terminology.  
  
    \item \textbf{OpenAI text-embedding-3-small.} This model produces dense vectors with 1,536 dimensions \cite{openai_embeddings}. It provides a computationally more efficient alternative to the large model.  
  
    \item \textbf{Hybrid GPT--BM25.} This approach first prompts gpt-5.4 to identify the $K$ CPT codes most relevant to the user query. The generated codes are then checked for validity. If fewer than $K$ valid codes are produced, the remaining candidates are supplemented using BM25 retrieval.   
\end{itemize}    
By comparing sparse, dense, and hybrid retrieval strategies we assessed the relative effectiveness of lexical matching and semantic similarity in identifying relevant price transparency records.  
  
\subsection*{Data preprocessing}  
  
\begin{figure}
    \centering
    \includegraphics[width=0.95\linewidth]{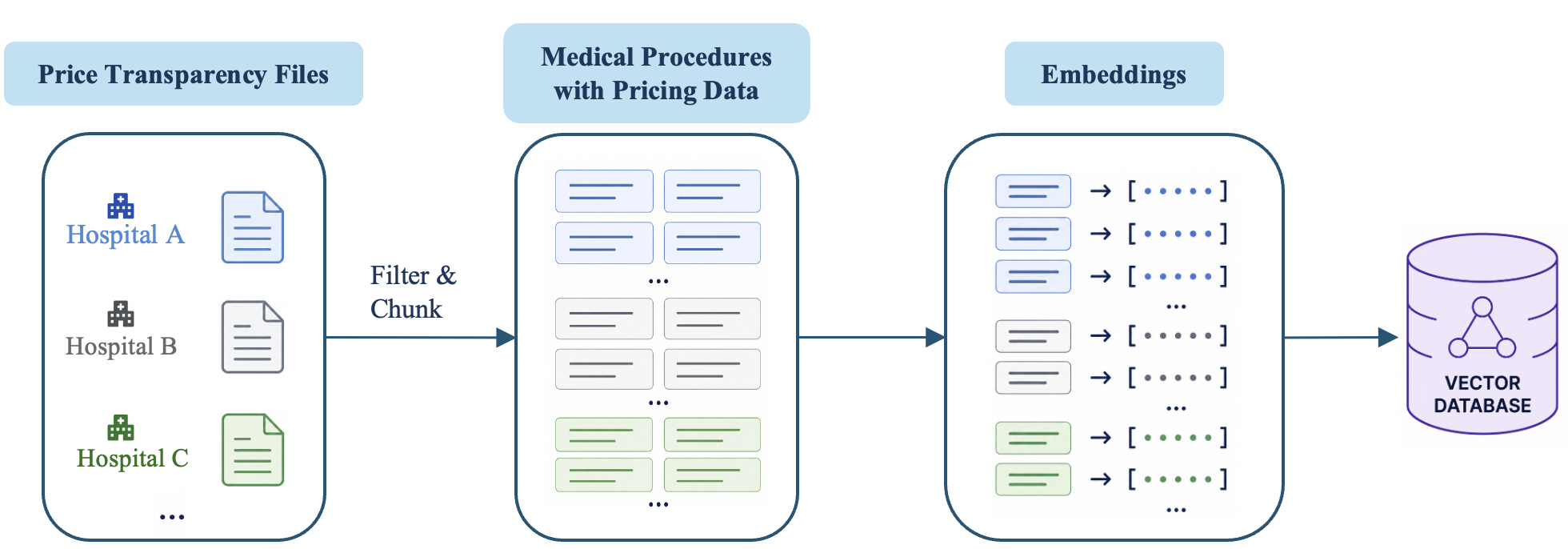}
    \caption{Preprocessing pipeline for price transparency files. Files are first filtered and processed so that each searchable chunk corresponds to a single CPT procedure. The long form service descriptions are then embedded and stored in a vector database for retrieval.}
    \label{fig:preprocessing}
\end{figure}

The PT files from different health systems are often poorly organized, structurally heterogeneous, and contain unclear or  redundant information. We therefore preprocessed the files as shown in Figure~\ref{fig:preprocessing} to standardize them for indexing and retrieval.

In preprocessing, we first excluded records without an associated Current Procedural Terminology (CPT) code \cite{dotson2013cpt}, since outpatient services are commonly billed using CPT codes. We then removed Medicare-related payment amounts and overlapping fee-schedule entries associated with research, administrative, or other noncomparable records. When procedure appeared multiple times across fee schedules or payment sources, we retained a single representative record. Remaining duplicate entries were resolved by retaining the record with the most detailed service description; when description length was identical, we retained the first occurrence.  

PT file service descriptions are often abbreviated and difficult to interpret. For example, "elbow 2 vws" is used for "x-ray of elbow, 2 views". Because these descriptions are embedded for indexing and storage in the retrieval database, abbreviations can degrade representation quality and weaken downstream retrieval performance. To mitigate this, we augmented each retained CPT-coded record with its official long-form CPT description from the Unified Medical Language System (UMLS) \cite{umls2024}. The enriched text was then used for embedding, indexing, and retrieval. Each final retrieval unit, or chunk, comprised a single CPT-coded procedure, its augmented service description, and the associated pricing information.

\begin{table}[htbp]  
\centering  
\caption{Summary of CPT code filtering across three hospitals}  
\label{tab:cpt_filtering}  
\begin{tabular}{lccc}  
\hline  
\textbf{Hospital} & \textbf{Total Records} & \textbf{Records with CPT Codes} & \textbf{Unique CPT Records} \\  
\hline  
Hopkins & $31222$ & $6487$ & $1955$ \\  
UMaryland & $334322$ & $3774$ & $2236$ \\  
Medstar & $191168$ & $8794$ & $3550$ \\  
\hline  
\end{tabular}  
\end{table}

\subsection*{Evaluation tasks and metrics}  
  
We evaluated the framework across three scenarios of increasing complexity:  
\begin{enumerate}  
    \item Estimating the price of an individual medical procedure, for example when a patient wants to know the cost of a scheduled X-ray, blood test, or vaccination.  
    \item Estimating the combined price of multiple procedures associated with a clinical or diagnostic visit, for example when a patient wants to know the likely total cost of an annual preventive visit that includes laboratory tests and screening services, or when a clinician seeks an estimate for the full set of services typically ordered during the evaluation of a specific complaint.  
    \item Estimating the expected pre-visit cost of care using information available before an outpatient encounter, such as the reason for the visit, presenting symptoms, and relevant medical and social history, for example when a patient describes symptoms online and wants to know the likely cost of the upcoming visit before scheduling.  
\end{enumerate}  

All of the prompts used in evaluations and data preprocessing are reported in the Supplementary Information.

\subsubsection*{\textit{Price estimation for individual procedures}}  
To evaluate individual-procedure price retrieval, we compiled a set of 277 services from the Johns Hopkins Hospital online price estimator, including 195 laboratory services, 68 radiology services, and 14 evaluation and management (E\&M) services. The estimator provides natural-language service descriptions together with their associated CPT codes. We treated these CPT-coded procedures as the reference labels, and used the corresponding prices in the hospital price transparency files as the ground-truth price values. In rare cases where a service description corresponded to multiple CPT codes, we resolved the ambiguity by consulting descriptions from the UMaryland and MedStar, so that each query was ultimately associated with a single correct CPT-coded procedure.

The service descriptions listed on the website were used as natural-language queries. For each procedure description, we generated five query variants to reflect common ways users might seek pricing information. For example, for the procedure ``Vitamin D'' the query variants were:  
\begin{enumerate}  
    \item ``Vitamin D'';  
    \item ``What is the price of a Vitamin D test/procedure?'';  
    \item ``What is the cost of a Vitamin D test/procedure?'';  
    \item ``Vitamin D cost''; and  
    \item ``Price of Vitamin D.''  
\end{enumerate}  
The queries were normalized before being embedded or submitted to the sparse retrieval system (see Figure~\ref{fig:rag_framework}).  
  
We evaluated both retrieval and generative steps. Retrieval performance was assessed using \textbf{Recall@K}, defined as the proportion of relevant procedures included among the top-$K$ retrieved records. For a query with $Z$ relevant CPT-coded procedures, let $C_K$ denote the number of relevant procedures among the top-$K$ retrieved records. Recall@$K$ was calculated as  
  
\begin{equation}  
    \mathrm{Recall@}K = \frac{C_K}{Z}.  
\end{equation}  
  
For the generation step, an LLM received the query and the top-$K$ retrieved candidate records and was asked to identify the corresponding procedure and its associated price. We evaluated procedure identification \textbf{accuracy}, defined as the proportion of queries for which the LLM selected the correct CPT-coded procedure from the retrieved candidates. Price prediction performance was assessed using the absolute percentage error (APE) between the predicted and reference prices:   
\begin{equation}  
    \mathrm{APE} = \left|\frac{y-\hat{y}}{y}\right|,  
\end{equation}  
where $y$ is the reference price and $\hat{y}$ is the predicted price. We report the mean of this quantity across evaluation examples as the \textbf{Mean Absolute Percentage Error (MAPE)}.

\subsubsection*{\textit{Price estimation for groups of procedures}}  
The second task simulated patients seeking the expected cost of multiple services associated with a clinical or diagnostic visit. This task is more challenging than individual-procedure retrieval because the system must identify multiple relevant records and aggregate their prices accurately.  
  
Based on the individual-procedure results, we selected the retrieval depth that produced the best overall performance, denoted by $K_{\mathrm{op}}$. We then evaluated the framework on groups of CPT-coded procedures of varying sizes. To approximate the composition of outpatient encounters, each group contained one procedure sampled from the evaluation and management category and additional procedures sampled from the laboratory and imaging categories. This sampling strategy was intended to reflect common outpatient visits more closely than uniform random sampling across all procedures.  
  
For each procedure in a group, retrieval performance was measured using \textbf{Recall@$K_{\mathrm{op}}$}. We summarized group-level retrieval performance using average Recall@$K_{\mathrm{op}}$, computed across the procedures in each group. During generation, the LLM received the retrieved candidates for each query and selected the procedure that best matched the query. We evaluated procedure selection using two complementary metrics:  
\begin{enumerate}  
    \item \textbf{Group Level Accuracy}, defined as the proportion of groups for which all constituent procedures were identified correctly; and  
    \item \textbf{Procedure Level Accuracy}, defined as the average, across procedure groups, of the fraction of constituent procedures identified correctly in each group. 
\end{enumerate}    
The LLM also generated a price for each selected procedure. For each procedure group, we summed the predicted procedure-level prices to obtain the predicted total cost and compared it with the corresponding reference total cost using MAPE; we report the mean of MAPE across groups.  
  
\subsubsection*{\textit{Realistic pre-visit price estimation}}  
  
The third task evaluated realistic pre-visit cost-estimation. Unlike the first two tasks, which used explicit procedure-based queries, this setting used information that may be available before an outpatient encounter, including the patient's reason for seeking care, symptoms, and relevant medical and social history.  
  
Ideally, this task would be evaluated using prospective patients and the prices of their subsequent encounters. However, establishing such a dataset is difficult because it requires linking pre-visit patient information with clinical services and billed prices while preserving patient privacy. We therefore constructed a synthetic evaluation dataset using LLM-generated patient personas, an approach that has been used in prior work to simulate realistic patient interactions \cite{ferber2026towards, kyung2026patientsim}.   

We began with publicly available, de-identified clinical notes from MTSamples.com \cite{mtsamples}. An LLM screened notes from the \textit{general medicine}, \textit{orthopedic}, \textit{consult -- history and physical}, and \textit{emergency room reports} categories and retained those describing non-life-threatening outpatient conditions. For each retained note, the LLM generated a plausible set of outpatient procedures that could be billed during the associated encounter. 
  
\begin{figure}
    \centering
    \includegraphics[width=0.95\linewidth]{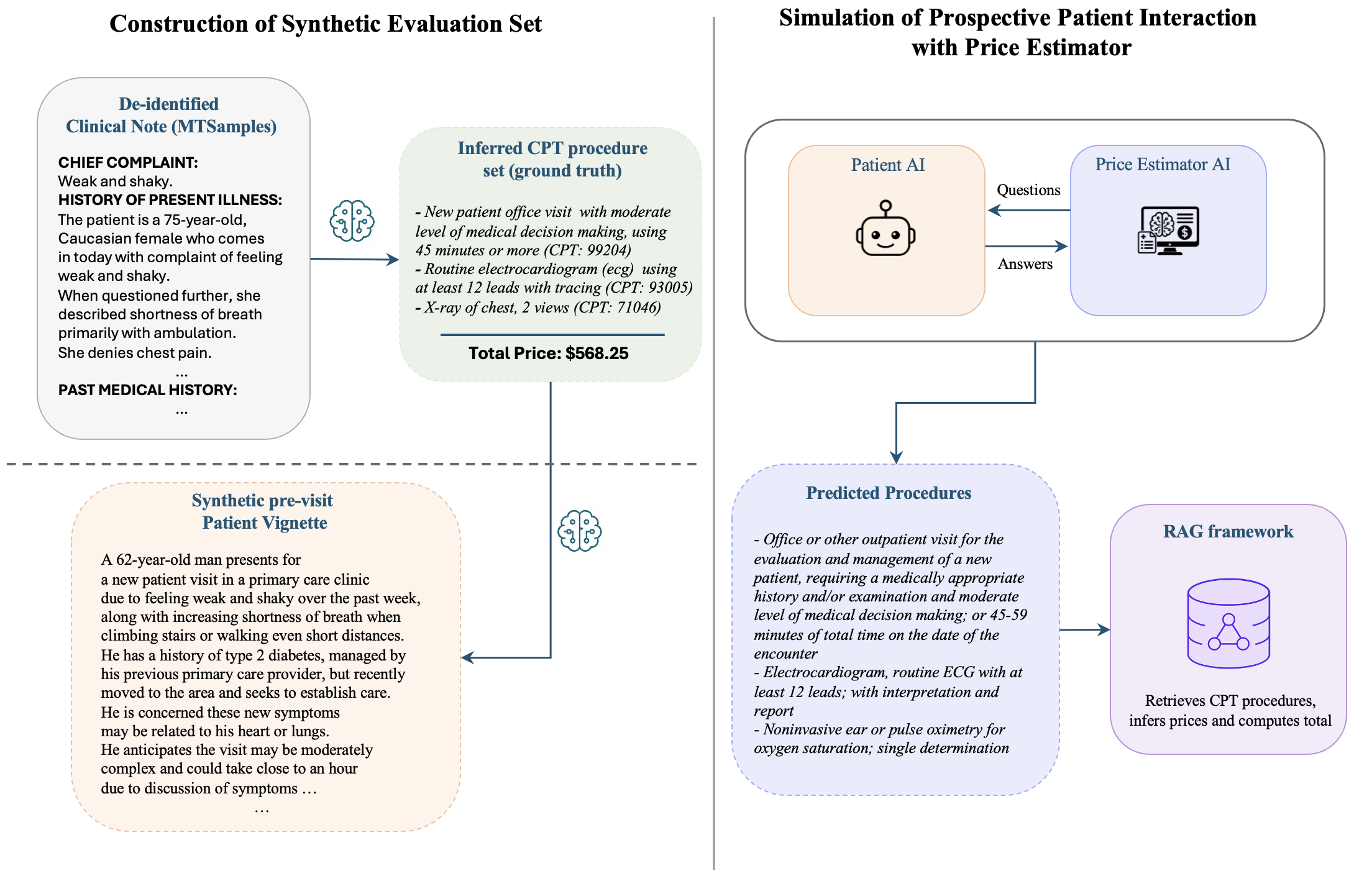}
    \caption{Synthetic evaluation framework for pre-visit cost estimation. De-identified clinical notes were used to generate reference procedure sets and corresponding total costs. Then an LLM playing role of synthetic patient persona interacted with the price-estimation system which predicted billed procedures and associated costs.}
    \label{fig:syntheticeval}
\end{figure}

We then used gpt-4.1 (version 04-14-2025) to generate a synthetic pre-visit patient vignette containing only information that would plausibly be available before the encounter, such as the chief concern, symptoms, and relevant history.  The number of profiles with given billed CPT group size is shown in Figure \ref{fig:profile_count_vs_billed_group_size}.  We intentionally used a model different from the one used in the main evaluation (gpt-5.4) so that the vignettes would be generated independently, thereby reducing the risk that the evaluation model could exploit artifacts or latent structure introduced during vignette generation. The procedure set derived from the original clinical note was treated as the reference set of billed procedures for that vignette, and the corresponding aggregate price served as the reference total cost (see Figure \ref{fig:syntheticeval}). 

For each vignette, an LLM acting as a prospective patient interacted with the price-estimation system which asked a set of questions about the chief complaint and relevant family, social, and medical history. Based on the information elicited during this interaction, the framework generated a predicted set of procedures and an associated total price. We compared the predicted procedure set with the reference procedure set and the predicted total price with the reference total price (Figure~\ref{fig:syntheticeval}).  

Performance was summarized using procedure-level recall, defined as the fraction of reference procedures correctly predicted, and MAPE between the predicted and reference total prices. So, both the system's ability to anticipate the services associated with an outpatient encounter and its ability to estimate the encounter's aggregate cost were assessed.

\section*{Results}
\label{sec:results}

\subsection*{Price estimation for individual procedures}
\begin{figure}  
    \centering  
    \includegraphics[width=0.95\linewidth]{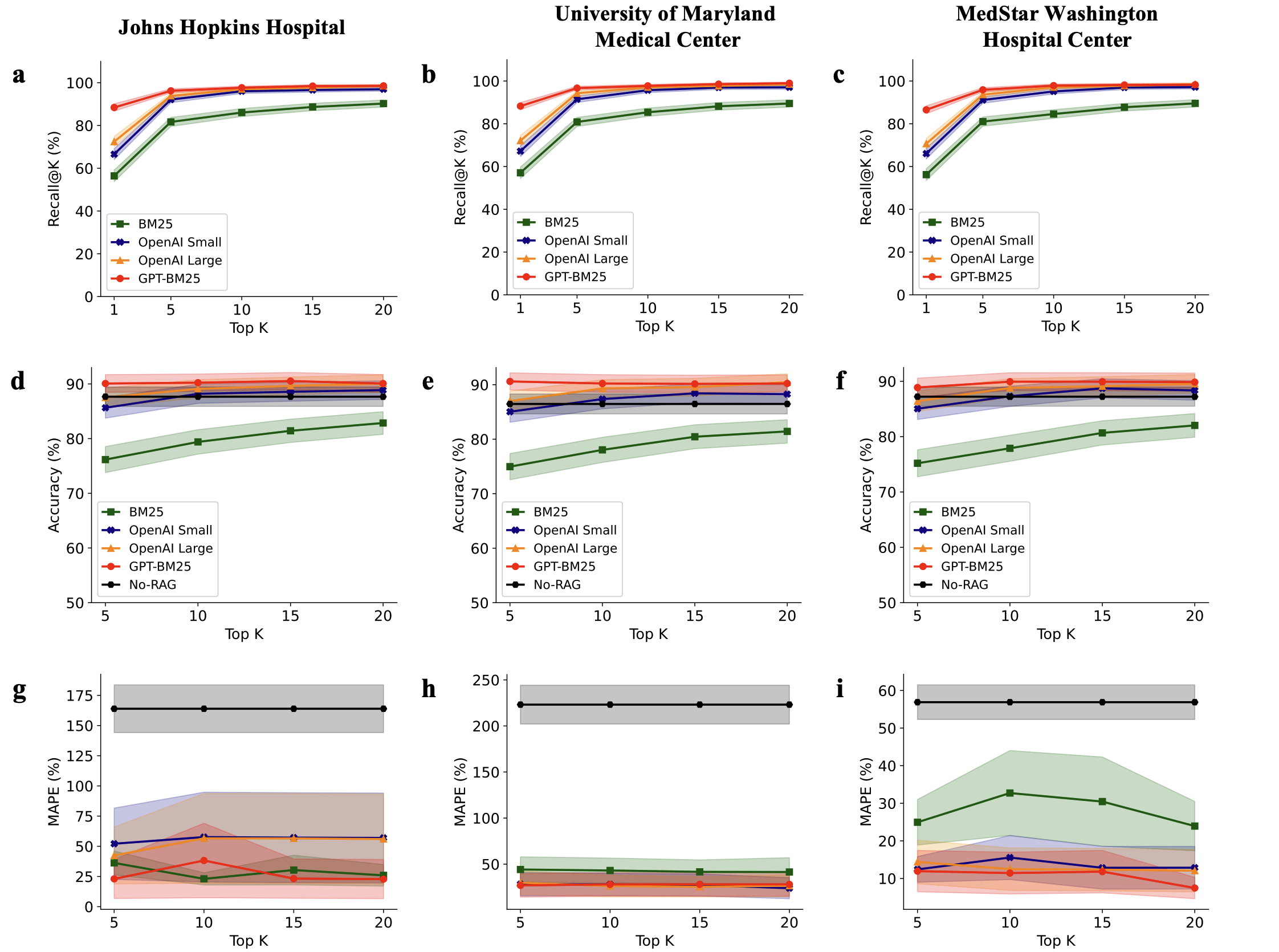}  
    \caption{Evaluation of price estimation for individual procedures across hospitals.   
    Retrieval scores are shown in panels (a--c), generative procedure-selection scores in panels (d--f),   
    and price MAPE in panels (g--i). Each column corresponds to a different hospital.}  
    \label{fig:singlequeryeval}  
\end{figure}

As expected, retrieval recall increased with the number of retrieved candidates, $K$   
(Figure~\ref{fig:singlequeryeval}a--c). Dense retrieval using OpenAI embeddings and the   
hybrid GPT--BM25 approach consistently outperformed sparse BM25 retrieval across hospitals.   
However, the benefits of increasing $K$ diminished beyond approximately $K=10$, at which   
point retrieval recall approached 98\%.  
  
Given the retrieved top-$K$ candidates, an LLM was tasked with identifying the procedure   
corresponding to the query and estimating its associated price. Procedure selection accuracy   
was generally higher when the candidate set was generated using either OpenAI embeddings or   
the hybrid GPT--BM25 retriever than when it was generated using BM25 alone   
(Figure~\ref{fig:singlequeryeval}d--f). The hybrid GPT--BM25 approach yielded the strongest   
overall performance, with a modest advantage over the embedding-based retrievers. Consistent   
with these per-hospital results, the cross-hospital averages in   
Table~\ref{tab:acc_mape_side_by_side} show that the LLM achieved its highest procedure-selection   
accuracy when conditioned on candidates retrieved by GPT--BM25.  
  
Price estimation errors varied substantially across hospitals   
(Figure~\ref{fig:singlequeryeval}g--i), likely reflecting differences in institutional pricing   
structures. Nevertheless, an LLM consistently provided the most accurate price estimates for GPT--BM25 retrievals,   
achieving a MAPE of approximately 20\% across hospitals   
(Table~\ref{tab:acc_mape_side_by_side}). We compared RAG with a no-retrieval baseline, in which the LLM was asked to infer both the procedure and its price from the query using only its parametric   
knowledge. Although the accuracy of procedure selection in this setting was comparable to that of the   
strongest retrieval systems, its price estimates were substantially less accurate. The no-retrieval baseline produced average MAPE of 147.98\%, considerably higher than that of even the   
BM25-based performance (Table~\ref{tab:acc_mape_side_by_side};   Figure~\ref{fig:singlequeryeval}d--i). Hence, response grounding is essential for accurate hospital-specific price estimation.

\begin{table*}[ht]  
\centering  
\begin{tabular}{lcccccccc}  
\toprule  
& \multicolumn{4}{c}{\textbf{Accuracy (\%)}} & \multicolumn{4}{c}{\textbf{MAPE (\%)}} \\  
\cmidrule(lr){2-5} \cmidrule(lr){6-9}  
\textbf{Retriever} & \textbf{K = 5} & \textbf{K = 10} & \textbf{K = 15} & \textbf{K = 20} & \textbf{K = 5} & \textbf{K = 10} & \textbf{K = 15} & \textbf{K = 20} \\  
\midrule  
BM25 & 75.44 & 78.45 & 80.85 & 82.11 & 35.12 & 32.95 & 34.15 & 30.45 \\  
OpenAI Small & 85.24 & 87.62 & 88.57 & 88.50 & 30.94 & 33.73 & 32.47 & 31.24 \\  
OpenAI Large & 86.97 & 89.07 & 89.42 & 90.03 & 28.86 & 31.81 & 31.44 & 31.55 \\  
GPT-BM25 & \textbf{89.85} & \textbf{90.13} & \textbf{90.20} & \textbf{90.05} & \textbf{20.63} & \textbf{25.90} & \textbf{21.05} & \textbf{19.44} \\  
No Retriever & 87.12 & 87.12 & 87.12 & 87.12 & 147.98 & 147.98 & 147.98 & 147.98 \\  
\bottomrule  
\end{tabular}  
\caption{Average generative selection accuracy and MAPE across hospitals for each retriever at different Top K values. Higher accuracy is better; lower MAPE is better.}  
\label{tab:acc_mape_side_by_side}  
\end{table*}  

\subsection*{Price estimation for groups of procedures}
\begin{figure}
    \centering
    \includegraphics[width=0.95\linewidth]{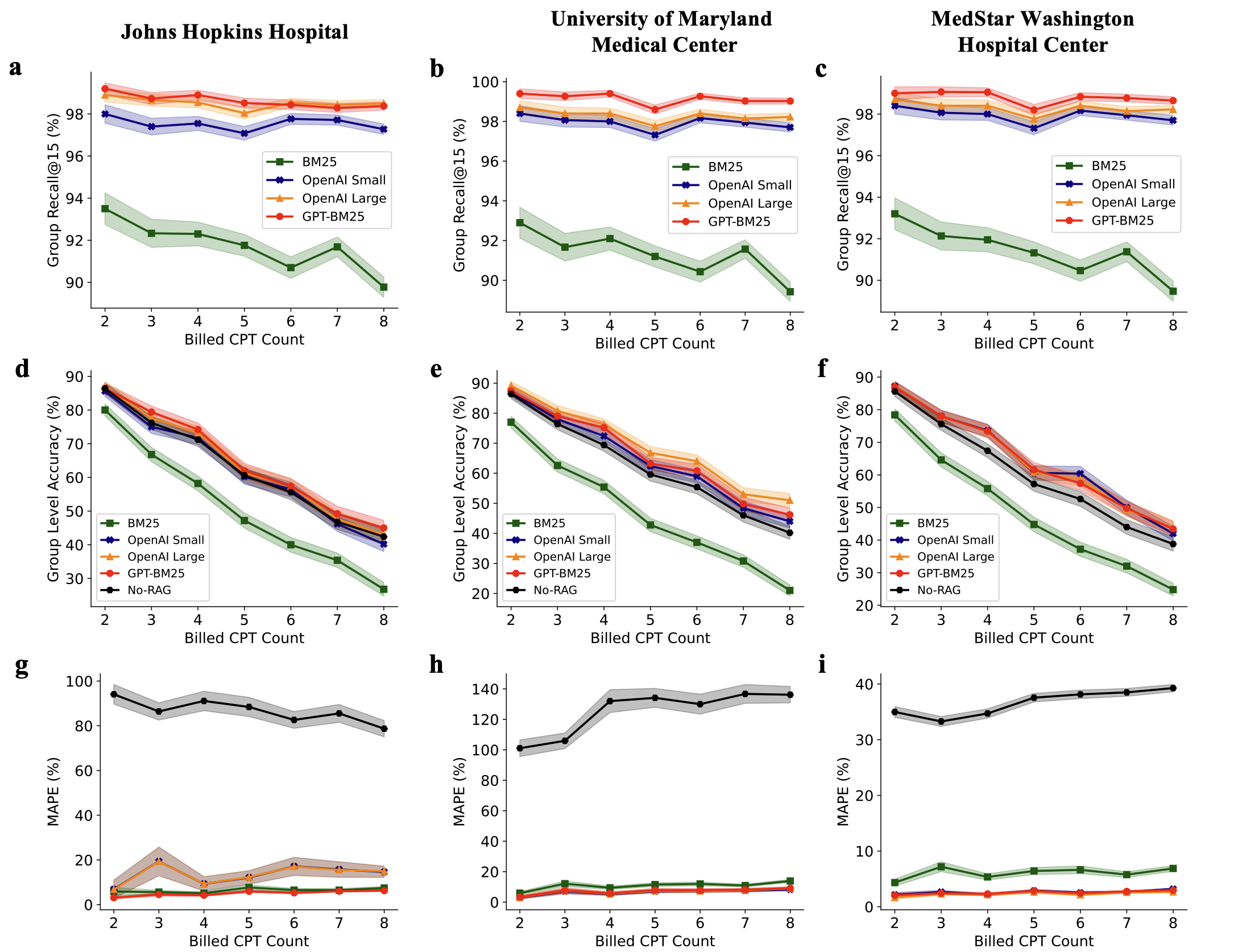}
    \caption{Evaluation of retrieval and price-estimation performance for groups of procedures across hospitals and group sizes.  Retrieval recall (a--c),  group level selection accuracy (d--f), and  MAPE of total cost of billed groups of procedures (g--i).}
    \label{fig:multiplequeryeval}
\end{figure}

Based on the individual procedure results, we selected $K_{\mathrm{op}}=15$ for the evaluation of retrieval and generative performance for queries on groups of procedures. GPT--BM25 achieved the strongest retrieval performance across hospitals and group sizes, with average recall of approximately 99\% for all group sizes (Figure~\ref{fig:multiplequeryeval}a--c). The OpenAI embedding-based retrievers performed comparably, whereas sparse BM25 consistently yielded the lowest retrieval recall.  
  
After retrievals, for each query, LLM selected the procedure that best matched procedure from the corresponding retrieved candidates. As expected, group level accuracy decreased as the number of procedures in a group increased, because correctly identifying every constituent procedure becomes progressively more difficult for larger groups (Figure~\ref{fig:multiplequeryeval}d--f). Selection performance was similar for GPT--BM25 and the OpenAI embedding-based retrievers. 

Notably, the no-RAG baseline achieved group level selection accuracies comparable to those of the RAG systems (Figure~\ref{fig:multiplequeryeval}d--f). However, this apparent similarity in selection performance did not translate into accurate total cost estimates: the no retrieval baseline produced substantially larger price-estimation errors (Figure~\ref{fig:multiplequeryeval}g--i).  
  
Across hospitals and group sizes, all retrieval-based methods achieved average total cost MAPE below 8\%. GPT--BM25 performed best, with an average MAPE of approximately 4.9\% (Table~\ref{tab:billed_group_average_performance}). The no retrieval baseline, in contrast, yielded a MAPE of 82.82\%. Low total-cost errors of our RAG system despite the decrease in group level accuracy for larger groups are explained by the higher procedure level accuracy. Although the LLM did not always identify all of the procedures in a group correctly, it selected the correct procedure for the majority of individual queries and therefore had access to the appropriate prices for most components of the total cost. 

\begin{table}[ht]  
\centering  
\begin{tabular}{lccc}  
\hline  
\textbf{Retriever} &  
\textbf{MAPE (\%)}  &  
\textbf{Group Level Accuracy (\%)}  &  
\textbf{Procedure Level Accuracy (\%)}  \\  
\hline  
BM25  
& $7.76 \pm 0.47$  
& $48.50 \pm 6.68$  
& $86.47 \pm 0.60$ \\  
  
OpenAI Small  
& $7.56 \pm 0.72$  
& $63.73 \pm 5.57$  
& $92.54 \pm 0.33$ \\  
  
OpenAI Large  
& $7.55 \pm 0.73$  
& $\mathbf{65.70 \pm 5.36}$  
& $\mathbf{93.18 \pm 0.32}$ \\  
  
GPT-BM25
& $\mathbf{4.90 \pm 0.39}$  
& $65.04 \pm 5.44$  
& $92.80 \pm 0.33$ \\  
  
No Retriever  
& $82.82 \pm 1.70$  
& $61.61 \pm 5.77$  
& $91.80 \pm 0.37$ \\  
  
\hline  
\end{tabular} 
\caption{Average generative performance across billing group sizes. Values are reported as mean $\pm$ standard error of the mean.}  
\label{tab:billed_group_average_performance} 
\end{table}  

\subsection*{Realistic pre-visit price estimation}
\begin{figure}  
    \centering  
    \includegraphics[width=0.95\linewidth]{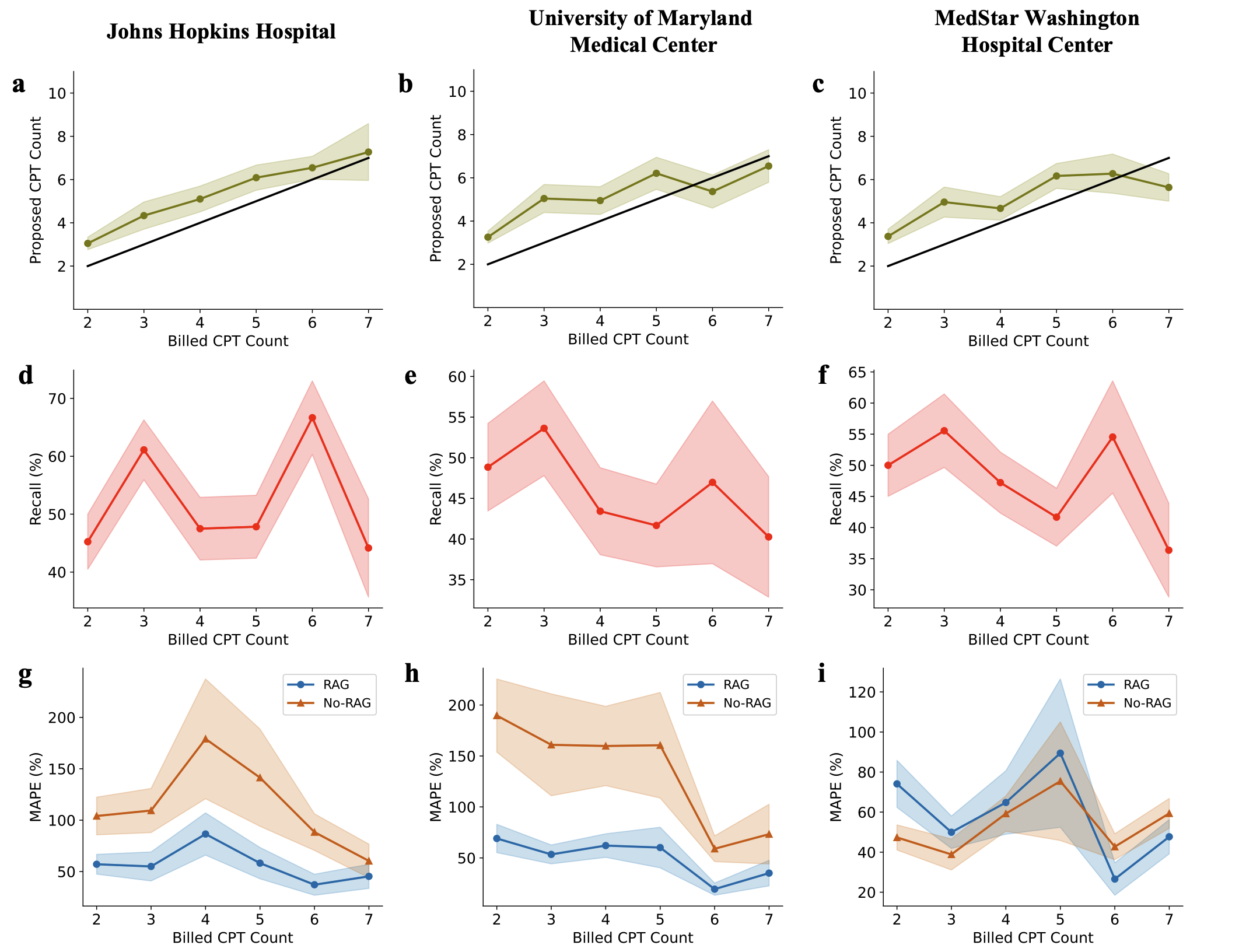}  
    \caption{Performance in the realistic pre-visit price-estimation evaluation.   
    Panels (a--c) compare the number of procedures proposed by the system with the number of reference billed procedures. Solid black line represents Y = X.   
    Panels (d--f) show procedure-level recall, and panels (g--i) show MAPE.   
    Each column corresponds to one hospital.}  
    \label{fig:realistic_visit_eval}  
\end{figure}   
In the synthetic pre-visit evaluation, for all three hospitals, the number of procedures proposed from the pre-visit information slightly exceeded the number of procedures ultimately designated as billed. Nevertheless, the two quantities followed a similar pattern: encounters with more reference billed procedures also had more procedures proposed by our framework (Figure~\ref{fig:realistic_visit_eval}a--c).  
  
We selected the hybrid GPT--BM25 retriever with $K=15$ for evaluation as it was the model with lowest errors for multiple procedure queries. In our framework, for each procedure proposed during the synthetic-patient interaction we retrieved corresponding candidate records from the relevant hospital's price transparency file. Then an LLM selected the procedure and associated price that best matched the proposed service. Across encounters with different numbers of reference billed procedures, approximately 50\% of the reference procedures were correctly selected to generate the price information (Figure~\ref{fig:realistic_visit_eval}d--f).  
  
We further compared aggregate price estimation to an LLM-only baseline without RAG. While for MedStar the RAG and no-RAG approaches produced comparable MAPE's across encounters with different numbers of reference billed procedures, for Hopkins and UMaryland, the RAG-based framework yielded lower errors than the LLM-only baseline (Figure~\ref{fig:realistic_visit_eval}g--i). Grounding price estimates in hospital-specific transparency files was on par or improved price estimation accuracies although large errors remain.

\section*{Discussion}
\label{sec:discussion}

We investigated the use of LLMs and RAG to improve hospital price transparency by developing a framework that analyzes hospital PT files and generates responses grounded in official pricing data. Although LLMs possess broad domain knowledge and demonstrate strong performance across scientific and medical reasoning tasks, they can produce hallucinations (factually inaccurate or unsupported statements) \cite{zhu2025can} that may have serious consequences when used for making medical or financial decisions. RAG architectures was shown to reduce hallucinations and improve performance by grounding model responses in retrieved evidence across medical domains \cite{xia2025mmed, li2026optimizing, liu2025improving}; however, its effectiveness depends critically on the quality of the retrievals. Because hospital pricing lies at the intersection of healthcare and financial information, reliable access to source grounded cost estimates is particularly important. We therefore compared the proposed RAG-based approach with an LLM-only baseline across three use cases: querying the prices of individual procedures, determining the costs of procedure groups, and estimating the total cost of prospective visits.

Across these use cases, the results show how retrieval quality and grounding contributed to price-estimation performance. Dense and hybrid retrieval provided more useful candidate sets than BM25 alone, with recall approaching 98\% at around 10 candidates, and GPT–BM25 achieved the strongest overall procedure-selection and price-estimation performance. Grounding was especially important for price accuracy: the no-retrieval baseline had a MAPE of 147.98\% for individual procedures, compared with approximately 20\% for GPT–BM25, with several high-error cases visible in the scatter plot (see Figure \ref{fig:scatter_price_gen_vs_true}).  This advantage extended to groups of procedures, where all retrieval-based methods achieved total-cost MAPE below 8\%, and GPT–BM25 achieved the lowest error at 4.90\%, compared with 82.82\% for the no-retrieval baseline. Although group-level selection accuracy declined as group size increased, procedure-level accuracy remained high, suggesting that identifying most but not necessarily all individual procedures can still support accurate cost estimates.

In the more realistic pre-visit evaluation, our framework correctly identified about half of the reference billed procedures, underscoring the difficulty of inferring billed procedures from medical history and highlighting procedure identification as a key area for improvement. Grounding estimates in price files reduced errors relative to the LLM-only baseline at Hopkins and UMaryland, but not at MedStar, where the two approaches performed similarly. Notably, the LLM-only baseline achieved substantially lower MAPE at MedStar than at Hopkins and UMaryland in both the individual- and multiple-procedure evaluations, despite high procedure selection accuracies. In the pre-visit simulations, however, only about half of the reference procedures were correctly identified on average. This may help explain the limited advantage of RAG at MedStar: when predicted procedures do not match the services ultimately billed, access to accurate hospital prices offers little benefit. Improving identification of likely billed procedures could therefore strengthen price estimates in realistic pre-visit settings

\subsection*{Limitations and Future Work}
This study has several limitations. First, the framework was evaluated using pricing data from three hospitals, which may limit the generalizability of the findings across institutions and healthcare markets. Evaluation at a broader range of hospitals is needed to better assess the performance. There are limitations due to contents of PT files, they often list multiple charges for the same procedure without clearly explaining the differences. We filtered these entries for our analysis, but the ambiguity may affect the accuracy of cost estimates. Requiring hospitals to explain such price variations could improve transparency and support more reliable estimates. Also, our estimates did not incorporate patients’ insurance coverage or negotiated rates. Although these data are difficult to obtain, they are necessary for estimating patients’ expected out-of-pocket costs. Finally, the synthetic pre-visit simulations may not fully reflect real-world use. A more realistic evaluation would require linking sensitive patient health information with billing records and insurance rates, which are difficult to access. The framework’s performance in actual clinical and billing workflows therefore remains to be established.

Future work should evaluate the framework across a broader range of hospitals and incorporate insurance data like negotiated rates, to produce estimates that better reflect patients’ expected out-of-pocket costs. Policy approaches that encourage or require hospitals to publish more insurance-specific pricing information should also be explored. We also plan to expand the framework to additional hospitals and develop a website for searching estimated case costs; a prototype is shown in Figure \ref{fig:prototype}.

\section*{Conclusion}
\label{sec:conclusion}

In summary, we developed a RAG framework to support outpatient cost estimation and price comparison. Our findings suggest that price transparency files can improve the precision of cost estimates for medical procedures and episodes of care. However, these evaluations remain preliminary. Broader validation across additional hospitals and real world settings, clearer characterization of price variation, and incorporation of insurance-specific pricing data will be essential to improve accuracy and generalizability. With further refinement, this approach could make hospital price transparency data more accessible, interpretable, and actionable for patients considering outpatient care.



\section*{Author contributions}

Kurmanbek Kaiyrbekov (Conceptualization, Data curation, Investigation, Methodology, Software, Visualization, Writing—original draft, Writing—review \& editing), Nicholas J. Dobbins (Conceptualization, Investigation, Methodology, Supervision, Writing—review \& editing), and Sean D. Mooney (Funding acquisition, Supervision, Writing—review \& editing)

\section*{Funding}
This work was supported by the Intramural Research Program of the National Human Genome Research Institute, National Institutes of Health.

\section*{Disclaimer}
This research was supported [in part] by the Intramural Research Program of the National Institutes of Health (NIH). The contributions of the NIH author(s) are considered Works of the United States Government. The findings and conclusions presented in this paper are those of the author(s) and do not necessarily reflect the views of the NIH or the U.S. Department of Health and Human Services.

\bibliography{main}

\pagebreak
\section*{Supplementary Appendix}
\label{sec:appendix}

\renewcommand{\thefigure}{S\arabic{figure}}
\setcounter{figure}{0}
\renewcommand{\thetable}{S\arabic{table}}
\setcounter{table}{0}

\begin{figure}[h!]  
\centering  
\setlength{\fboxsep}{10pt}  
\fbox{  
    \parbox{0.95\textwidth}{ 
        {\fontfamily{pcr}\selectfont\footnotesize

        You are a medical coding assistant specializing in CPT-style descriptions.  
        
        Your task is to rewrite a user question/query into the matching CPT procedure description written in clear, standardized language.

        When the user provides a query generate only a long-form descriptor that is as close as possible in meaning and style to the official CPT long description.
        
        \# Rules For Generating the CPT Procedure Description
          
            * Return only a single sentence CPT procedure description.
            
            * Do NOT include newlines at the end of the description.
            
            * Output should be English language only.  
            
            * Do NOT include CPT codes, explanations or anything else.  
            
            * Preserve the original clinical intent and scope of the procedure.    
            
            * Do NOT add procedures, tests, panels, or clinical details not implied by the query/question.    
            
            * Be as specific as possible when describing the procedure and avoid generalizing via bundled procedures.
            
            * If the original query contains a precise and closely matching CPT description, you can return same or very similar description. 
            
            * For office visits, if the query does not explicitly mention established patient, you can assume it is a new patient visit. 
        
        \# Output Format Requirements 

            * Use lowercase letters only.  
            
            * Use digits for numbers where applicable. 

        \# Examples of Query and Desired CPT Procedure 
        
       Query: What is the cost of mri upper extrermity non joint wo\&w contrast test/procedure?  
        
        CPT Description: mri scan of arm before and after contrast  
        
        Query: Cost for an mam xray surgical specimen test/procedure?  
        
        CPT Description: x-ray of surgical specimen 
        
        Query: What is the cost of ct colonography screening test/procedure?
        
        CPT Description: screening ct scan of large intestine
        
        Query: What is the price of renin test/procedure? 
        
        CPT Description: renin (kidney enzyme) level

        Query: What is the price of thiamine test/procedure? 
        
        CPT Description: vitamin b-1 (thiamine) level

        Query: What is the price of potassium urine test/procedure? 
        
        CPT Description: urine potassium level

        Query: What is the price of urea nitrogen urine test/procedure?
        
        CPT Description: urea nitrogen level to assess kidney function, urine

        Query: What is the price of initial new patient preventive medicine evaluation (18-39 years) test/procedure?
        
        CPT Description: initial new patient preventive medicine evaluation (18-39 years)
        
        Query: What is the cost of amniocentesis test/procedure??
        
        CPT Description: removal of sample of amniotic fluid surrounding fetus for diagnosis
    
}
    }  
}  
\caption{Prompt to rewriting queries (normalization).}  
\label{fig:query_normalization}  
\end{figure}

\begin{figure}[h!]  
\centering  
\setlength{\fboxsep}{10pt}  
\fbox{  
    \parbox{0.95\textwidth}{ 
        {\fontfamily{pcr}\selectfont\small
        You are a medical billing coding assistant helper that selects CPT procedure(s) that best match a given description or query. Your task does NOT involve medical advice, diagnosis, treatment recommendations, or patient guidance. This task is strictly administrative: mapping procedure description or query to candidate CPT codes and their associated gross charges. Given a procedure description or query, select only the CPT code(s) from the provided candidates that correctly match the procedure or query and their corresponding prices. Use gross charges as procedure prices. Do not invent CPT codes or prices; only select from the provided candidates. If multiple candidates are plausible, prefer the procedure that most closely matches the stated clinical scope and is not more extensive or bundled than necessary. If no candidate is an exact match, select the closest reasonable match in clinical scope.

        Match the following procedure description or query to the most appropriate CPT
code(s) from the candidate list.

Procedure description or query (clinical context):

\{query\}

Candidate CPT codes with prices:

\{candidates with pricing info\}

Select the minimum number of CPT codes required to represent the procedure.

Return ONLY the matching CPT code(s) and prices as a JSON array of objects with keys cpt\_code (string) and price (number). Do not include explanatory text.
    
}
    }  
}  
\caption{Prompt to select CPT-coded procedures and associated prices from retrieved candidates for a given query.}  
\label{fig:selection_from_candidates}  
\end{figure}

\begin{figure}[h!]  
\centering  
\setlength{\fboxsep}{10pt}  
\fbox{  
    \parbox{0.95\textwidth}{ 
        {\fontfamily{pcr}\selectfont\small

You are a medical billing coding assistant helper that selects CPT procedure(s) that best match a given description or query. Your task does NOT involve medical advice, diagnosis, treatment recommendations, or patient guidance. This task is strictly administrative: mapping procedure description or query to existing CPT codes nd their associated gross charges. Given a procedure description or query, generate only the CPT code(s) that correctly match the procedure or query and their corresponding prices. Use gross charges as procedure prices from price transparency files of \{hospital\_name\} published July 1, 2025.

The following is a medical billing classification task. All procedure
descriptions, queries, and outputs are only for research purposes.

Find the most appropriate CPT code(s) and price(s) for the procedure/query below.

Procedure description or query (clinical context):

\{query\}

Generate the minimum number of CPT codes required to represent the procedure.

Return ONLY the matching CPT code(s) and prices as a JSON array of objects with keys cpt\_code (string) and price (number). Do not include explanatory text.
    
}
    }  
}  
\caption{Prompt to directly generate cpt procedures and associated prices without retrievals. (No-RAG case)}  
\label{fig:direct_cpt_price_generation}  
\end{figure}

\begin{figure}[h!]  
\centering  
\setlength{\fboxsep}{10pt}  
\fbox{  
    \parbox{0.95\textwidth}{ 
        {\fontfamily{pcr}\selectfont\small 
    You are a clinical visit classifier for healthcare research.  
    
    Task:  
    Classify a clinical note as YES or NO.  
    
    Return YES only if ALL of the following are true:  
    
    1. The note describes a patient visit in a non-life-threatening outpatient or ambulatory setting, such as:  
        
        - office visit  
        
        - clinic visit  
        
        - consultation  
        
        - emergency department / emergency room visit that is not critical or life-threatening  
   
    2. The visit implies that at least one laboratory test and/or imaging study will be billed after the visit.  
    
    3. No surgery or operative procedure was performed during the visit.  
    
    4. The note reflects an actual evaluation/management visit, not only a standalone procedure, discharge summary, or administrative documentation.  
    
    Return NO if ANY of the following are true:  
    
    - The visit is inpatient, ICU, operative, perioperative, autopsy, or discharge-only.
    
    - The reason of visit is major trauma, trauma activation, or another clearly life-threatening traumatic injury.  
    
    - The note describes a procedure without a clear visit/evaluation context.  
    
    - Surgery or an operative intervention was performed during the visit.  
    
    - The presentation is life-threatening, critical, or unstable, including but not limited to:  
        
        - major trauma  
        
        - stroke  
        
        - myocardial infarction / STEMI / NSTEMI  
        
        - sepsis or septic shock  
        
        - respiratory failure  
        
        - shock  
        
        - cardiac arrest  
        
        - altered mental status with instability  
        
        - severe hemorrhage  
        
        - another clearly critical condition  
    
    - The note does not clearly support that labs and/or imaging would be ordered after the visit.  
    
    Decision rules:  
    
    - If there is ambiguity, uncertainty, or insufficient evidence, return NO.  
    
    - Be conservative: YES requires clear support for every required criterion.  
    
    - Emergency room visits are YES only if they are clearly non-critical and meet all other criteria.  
    
    Output format:  
    
    Return only one word:  
    YES  
    or  
    NO
}
    }  
}  
\caption{Prompt for filtering notes from MTSamples to obtain  outpatient or non-life-threatening emergency department visits.}  
\label{fig:filtering_mts}  
\end{figure}

\begin{figure}[h!]  
\centering  
\setlength{\fboxsep}{10pt}  
\fbox{  
    \parbox{0.95\textwidth}{ 
        {\fontfamily{pcr}\selectfont\small 
    You are an expert medical coding assistant for healthcare research.

    Task:
    
    Given a clinical visit note, generate likely billed CPT codes with corresponding descriptions.

    Requirements:
    
    - Return exactly one most likely billed evaluation and management CPT code and description in e and m code.
    
       * The evaluation and management procedure description should include complexity level for emergency visits.
       
       * The evaluation and management procedure description should include complexity in terms of time needed for outpatient visits.
       
       * Infer complexity from the note.
       
    - In addition, return likely billed lab and/or radiology CPT codes and corresponding descriptions in lab or radiology codes.
       
       * For labs/radiology, include the most likely tests or imaging studies that are expected to be billed after the visit.
       
       * Do not include overlapping lab codes that would duplicate the same scope of testing or results. 
       
       * If similar tests are necessary their official descriptions should differentiate them.
       
       * Do not include duplicate tests.
    
    - Only include CPT codes that are clinically supported by the note.
    
    - Prefer commonly billed procedures by medical coders over rarely billed ones.
    
    - The procedure descriptions should be similar to official long form CPT descriptions.
    
    - Do not include ICD diagnosis codes.
    
    - Do not include surgery, anesthesia, pathology-only, inpatient, ICU, discharge, or procedure-only codes.
    
    - Be conservative. If a lab or imaging CPT is unclear, choose the most clinically plausible CPT based on the note.

    Output:
    
    Return only the structured answer with:
    
    - e and m code
    
    - lab or radiology codes
}
    }  
}  
\caption{Prompt for generating billed procedures from MTSamples notes.}  
\label{fig:billed_proc_gen_mtsapmles}  
\end{figure}

\begin{figure}[h!]  
\centering  
\setlength{\fboxsep}{10pt}  
\fbox{  
    \parbox{0.95\textwidth}{ 
        {\fontfamily{pcr}\selectfont\footnotesize 

You are generating a fictional patient pre-visit profile for healthcare research.  
    
    Task:  
    Given a short reason of visit and list of services billed to a patient after a single clinical encounter write very comprehensive and realistic patient's vignette BEFORE the visit. 
    
    Goal:  
    Create a plausible PRE-VISIT profile of a fictitious patient that naturally upon visit to clinic leads specifically to the billed services. 
    Avoid details that would strongly suggest additional unrelated services not included in the encounter.  
    
    Important Constraints:  
    
    - Focus on the patient's condition before visit that would warrant billed procedures
    
    - Do not mention anything related to post-visit assessment, treatment, outcomes or assessment during visit  
    
    - Do not mention billing codes  
    
    - Do not explicitly name or list the billed procedures or services  
    
    - Do not write a diagnosis unless it would already have been known before the visit  
    
    - The description should be specific enough that a clinician or medical coder could reasonably infer the likely billed services after interviewing the patient  
    
    - Keep the vignette internally consistent and clinically plausible 
    
    - If the listed services appear to reflect more than one concern, you may describe more than one presenting problem rather than forcing everything into a single symptom narrative  
    
    Write the output using exactly these sections:  
    
    Summary:  
    - Reason for seeking care 
    
    - Whether this is a new visit or follow-up, when relevant  
    
    - Visit setting when relevant (for example: primary care, specialist clinic, urgent care, emergency department)
    
    - Mention how long the visit is expected to take or complexity of visit from patient perspective, if relevant  
    
    Present symptoms:  
    
    - Main symptoms or concerns  
    
    - Onset, duration, progression, severity, and relevant context    
    
    Absent symptoms:  
    - Relevant symptoms the patient does not have  
    
    Recent tests/results: 
    
    - For follow up visits, include relevant normal or abnormal prior tests from an earlier visits when appropriate 
    
    - If helpful, include relevant normal or abnormal outside labs, imaging, or other prior test results obtained before this encounter    
    
    - Only include tests that support the vignette and makes it consistent with billed procedures
    
    - If a commonly expected test was not billed, use this section to mention if it was done previously    
    
    Past medical history:  
    
    - Relevant prior medical conditions, or "None" if not applicable
    
    - Include plausible prior evaluations or prior clinician visits if needed for coherence 
    
    Past surgical history: 
    
    - Relevant prior surgeries, or "None" if not applicable  
    
    Social history:  
    
    - Relevant social history such as tobacco, alcohol, drug use, occupation, living situation, sexual history, or other context when clinically relevant  
    
    - If not relevant, keep this brief  
    
    Output requirements: 
    - Return only the fictional patient profile.  
    - Do not include explanations, meta-commentary, or reasoning.  
}
    }  
}  
\caption{Prompt for generating synthetic vignettes.}  
\label{fig:profile_generation_from_mtsamples_w_gpt4.1}  
\end{figure}

\begin{figure}[h!]  
\centering  
\setlength{\fboxsep}{10pt}  
\fbox{  
    \parbox{0.95\textwidth}{ 
        {\fontfamily{pcr}\selectfont\small 
You are a prospective hospital patient using an AI-based price estimator before seeking care. 
    You will be given a synthetic patient profile and must answer as that patient.
    
    Here is the patient profile: 
    
   \{ profile description \}

    Requirements:
    
    - Be honest and strictly follow the synthetic profile and do not contradict it in your responses.
    
    - Be thorough and respond in first person, do not miss any relevant details from the profile.
    
    - Make sure to respond to each question.

}
    }  
}  
\caption{Prompt used for LLM to play a role of synthetic patient.}  
\label{fig:patient_actor_prompt}  
\end{figure}

\begin{figure}[h!]  
\centering  
\setlength{\fboxsep}{10pt}  
\fbox{  
    \parbox{0.95\textwidth}{ 
        {\fontfamily{pcr}\selectfont\small 
Please answer to the following questions:

    1. What department or care setting are you visiting and how long do you expect the visit to take or how complex is the visit from your perspective?
    
    2. Are you a new patient or is this a follow-up visit?
    
    3. What are your main symptoms and concerns, and how long have you been experiencing them?
    
    4. List symptoms you do not have that are relevant to your main concerns.
    
    5. What relevant prior medical history, surgical history, and recent test results do you have?
    
    6. What relevant social history, lifestyle factors, or other context is important for the doctor to know?
    
    7. Did you have any prior evaluations or test results that you would like to share?
    
    8. Is there any other information you would like to provide to the doctor before your visit?

}
    }  
}  
\caption{Questions asked to synthetic patient before predicting billed procedures.}  
\label{fig:ai_system_prompt}  
\end{figure}

\begin{figure}
    \centering
    \includegraphics[width=0.95\linewidth]{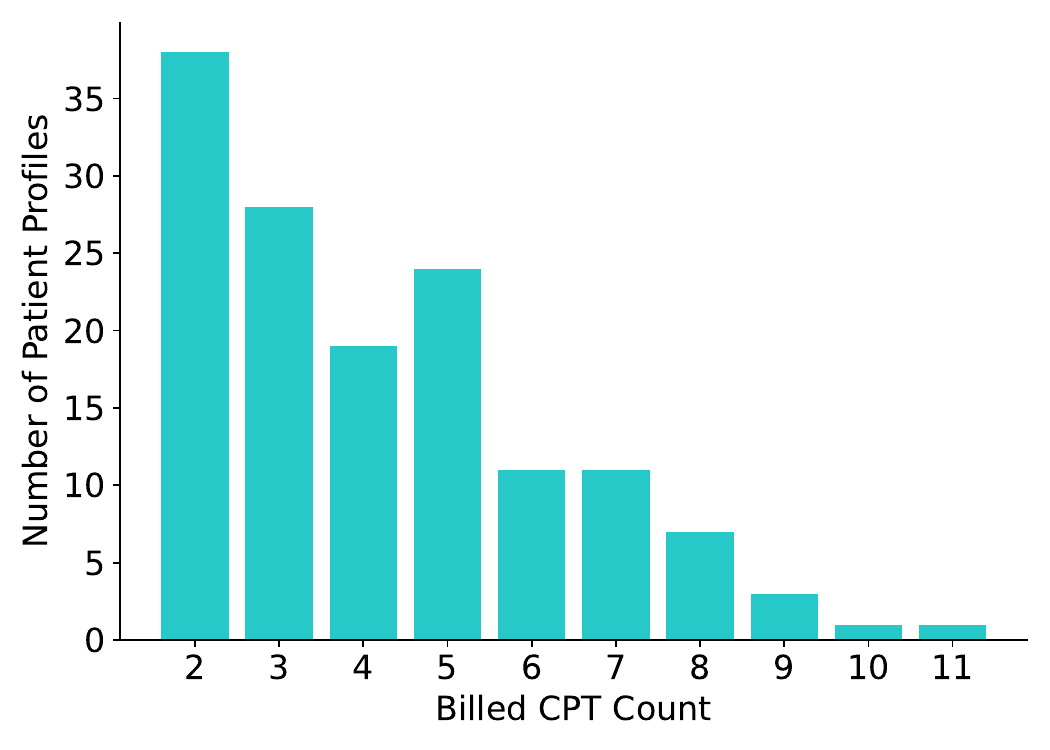}
    \caption{Number of patients vs count of billed CPT procedures generated by LLM. If number of profiles for billed CPT count were less than 10, they were excluded from final analysis.}
    \label{fig:profile_count_vs_billed_group_size}
\end{figure}

\begin{figure}
    \centering
    \includegraphics[width=0.95\linewidth]{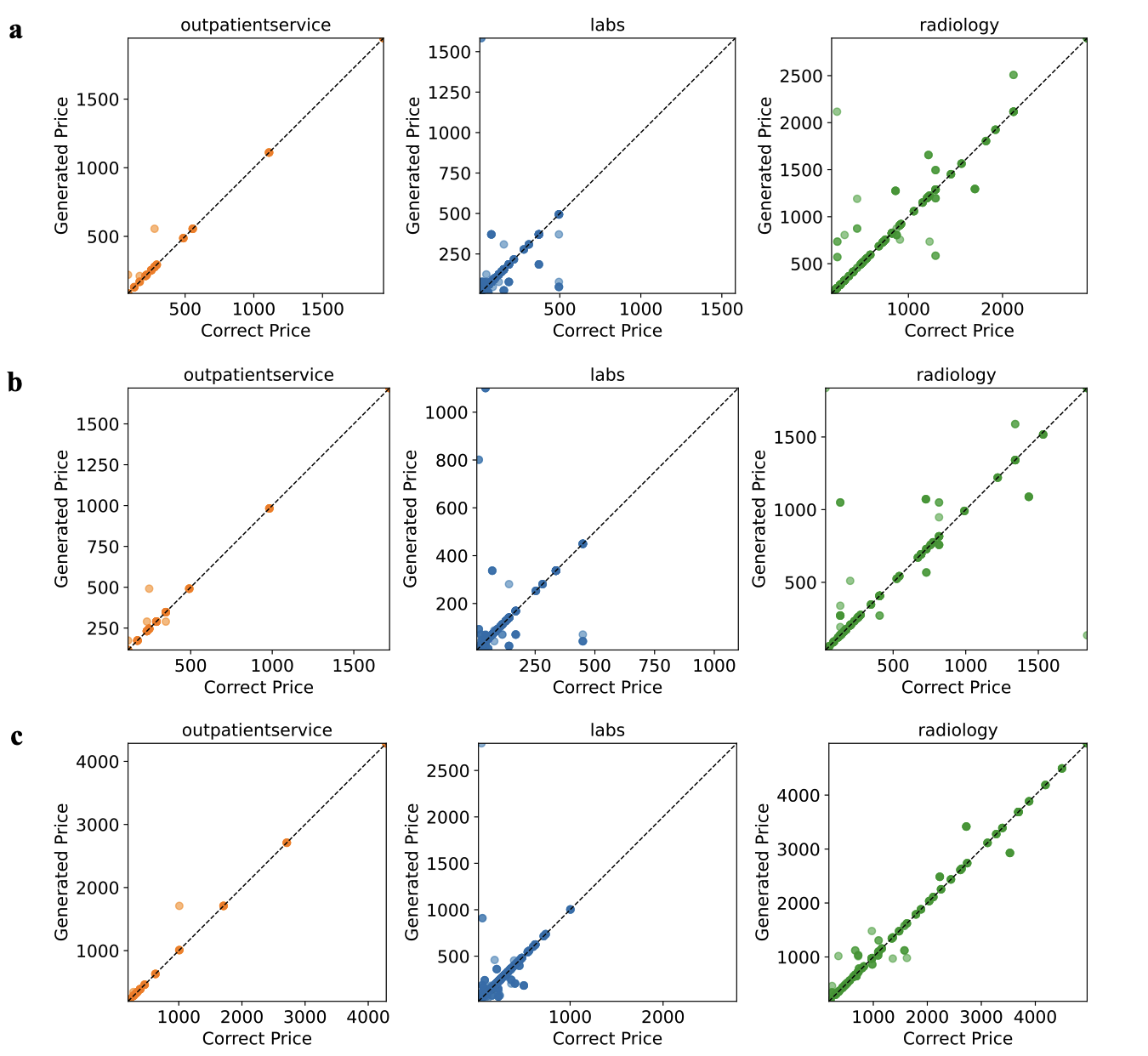}
    \caption{Scatter plots of Hopkins (a), UMaryland (b), and Medstar (c) for generated versus actual prices for different billed service types for GPT-BM25 retriever and Top K = 15 retrievals. }
    \label{fig:scatter_price_gen_vs_true}
\end{figure}

\begin{figure}
    \centering
    \includegraphics[width=0.95\linewidth]{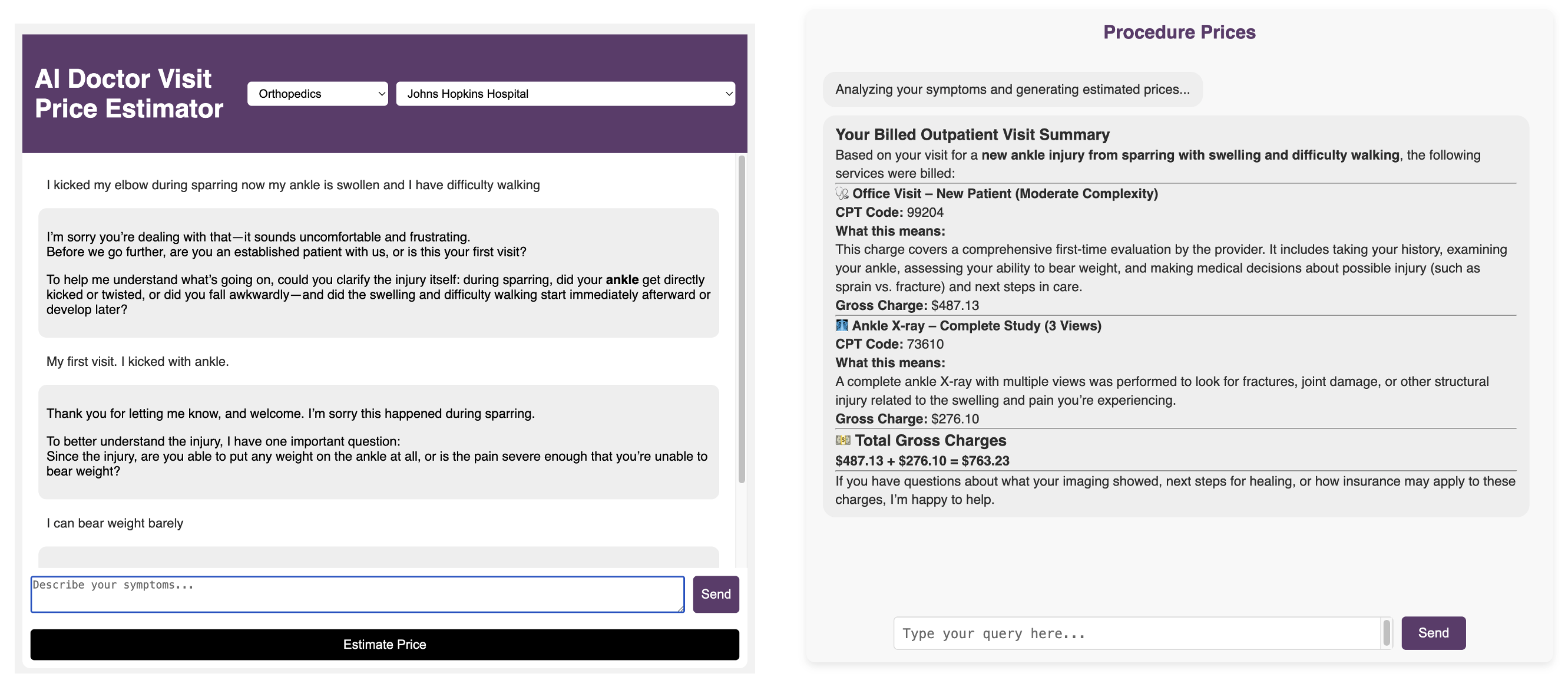}
    \caption{Current prototype of a website for estimating hospital care price.}
    \label{fig:prototype}
\end{figure}


\end{document}